# DOMINO: Domain-aware Loss for Deep Learning Calibration

Skylar E. Stolte[1], Kyle Volle[2], Aprinda Indahlastari[3,4], Alejandro Albizu[3,5], Adam J. Woods[3,4,5], Kevin Brink[6], Matthew Hale[2], and Ruogu Fang[1,3,7,*]

*[1]J. Crayton Pruitt Family Department of Biomedical Engineering, Herbert Wertheim College of Engineering, University of Florida (UF), USA*
*[2]Department of Mechanical and Aerospace Engineering, Herbert Wertheim College of Engineering, UF, USA*
*[3]Center for Cognitive Aging and Memory, McKnight Brain Institute, UF, USA*
*[4]Department of Clinical and Health Psychology, College of Public Health and Health Professions, UF, USA*
*[5]Department of Neuroscience, College of Medicine, UF, USA*
*[6]United States Air Force Research Laboratory, Eglin Air Force Base, Florida, USA*
*[7]Department of Electrical and Computer Engineering, Herbert Wertheim College of Engineering, UF, USA*

## Abstract
Deep learning has achieved the state-of-the-art performance across medical imaging tasks; however, model calibration is often not considered. Uncalibrated models are potentially dangerous in high-risk applications since the user does not know when they will fail. Therefore, this paper proposes a novel domain-aware loss function to calibrate deep learning models. The proposed loss function applies a class-wise penalty based on the similarity between classes within a given target domain. Thus, the approach improves the calibration while also ensuring that the model makes less risky errors even when incorrect. The code for this software is available at https://github.com/lab-smile/DOMINO.

[*]Corresponding author: ruogu.fang@ufl.edu

## 1. Introduction to Deep Learning Calibration

Modern deep learning models achieve spectacular accuracy; however, they often disregard calibration analysis [9]. Confidence calibration relates the model output prediction score ("the confidence") to the true likelihood of a class being correct [9, 2]. In other words, a calibrated model has an X percent chance of getting each data point that has X output probability correct. This is related to the model uncertainty, which corresponds to the classification noise [2]. Confidence calibration and uncertainty details are highly associated with model interpretability [2]. Hence, models with high accuracy and poor calibration could appear groundbreaking during research development, yet these same models may not be trustworthy for clinical deployment.

One reason for this poor translation to clinical data is that clinical data is often more variable than research data. Indeed, generalization to out-of-domain (OOD) data is a huge challenge in machine learning [18]. OOD data refers to any testing data that is different in distribution from the training data. In the research setting, it is easy to design an ideal testing dataset that matches our training data. This is performed by dividing one dataset into training and testing data splits. A fundamental problem with current deep learning models is that the high accuracy on such datasets often does not translate to OOD testing data. Different factors that may affect the distribution of clinical data include variability in data collection parameters, differences between patients, and rare data classes. For instance, changing to a different imaging scanner manufacturer may drastically lower segmentation performance [6, 15]. Rare data classes are at the highest risk of these changes [7]. Poor performance on rare data is potentially dangerous, as the rarest data in medicine are often in diseased states [7]. There is prior evidence that confidence calibration can help [18].

In addition, model calibration can improve a model's ability to detect when it is most likely to fail [4]. Failure prediction includes successfully detecting OOD data and reporting model confidence [14]. This feedback is equally as important as improving the performance because it strengthens a model's interpretability. For instance, a disease classification model could sort patient cases based on confidence rather than just the binary label. This information is useful in clinical screening since the true labels will not always be available. In this setting, a prediction with low confidence would be highlighted for clinician review. Overall, a calibrated machine has more potential for embedded safety features when compared to an uncalibrated model, even when the performance is equal.

Model calibration has the potential to improve failure detection, aid model interpretability, and possibly even improve generalizability. Yet, most modern deep learning models are poorly calibrated. Hence, DOMINO [17] was developed to calibrate deep learning models. The DOMINO loss function that computes penalties for incorrect classifications based on class-wise similarities. DOMINO computes the relevant class similarities using either task-driven or data-driven similarities. This approach improves performance by learning more true information about class representation, as opposed to traditional deep learning algorithms that fight class similarity. In addition, DOMINO can be adjusted so that classes that are safer to confuse are closer together. In this way, DOMINO loss allows a model to make safer and more meaningful mistakes when wrong. These features add to the overall safety and effectiveness of this calibration approach.

## 2. The DOMINO Methodology

DOMINO calibration is constructed as a loss regularization term for deep learning. The base loss function only relies on Python3 and any PyTorch version. CUDA is optimal for best performance, but it is not required. The DOMINO-regularized loss function is

$$L(y, \hat{y}) + \beta y^T W \hat{y} \qquad (1)$$

where $L$ is any loss function, $y$ is the one-hot encoded true label, and $\hat{y}$ is the softmax output. The parameter $\beta$ can take on any value between zero and one, and the entire novel loss term is scaled by $\beta$. The matrix $W$ represents a generic regularization term of size $N \times N$, where $N$ is the number of classes. The diagonals are zero, whereas the off-diagonals represent the penalties for confusing classes.

The overall loss function is simple yet efficient so that it can be dynamically altered for a given task. DOMINO can work with either classification or segmentation, whereas the segmentation approach is regarded as classification on a pixel-wise basis. However, one caveat of the loss function is that it is optimized for multiclass problems. The reason is that DOMINO operates on the fundamental logic that the functional performance of a model on a particular task may be improved by treating certain classes as being less risky to confuse than others. These ideas lose their meaning in a binary problem since in a binary case the two classes are maximally separated.

DOMINO is easy to implement and adjust to different deep learning frameworks because its overall structure is simple. Equation 1 operates according to basic matrix multiplication. At the same time, DOMINO is effective due to its adaptability across tasks. This is because the only parameter that is inherently task-specific is the $W$ term, and $W$ can be modified based on the specific user task. This term refers to a class-wise weighting penalty that regularizes our loss based on specific penalties for confusing a given class for any other class. The scheme does not need to be symmetric; for instance, a disease severity model could give step-wise increases in penalization such as giving higher weight to false negatives. Our prior works [17] use two main $W$ schemes that can be broadly categorized as machine-level confusion (confusion matrix method (DOMINO-CM)) or expert-guided groupings (hierarchical class method (DOMINO-HC)).

## 3. The Advantages of using DOMINO

Our prior experiments [17] show that DOMINO can improve both accuracy and calibration metrics across many classification and segmentation tasks. DOMINO is easy to implement in existing projects via a quick addition to the loss function. DOMINO is particularly advantageous for works in which certain mistakes would carry lower risks than others. For example, a theoretical deep learning model controls a self-driving car which has to swerve due to frozen roadways. There are three “objects” in the car’s vision: a pedestrian, a bicycle rider, and a stop sign. In this situation, an uncalibrated model might equally confuse a bicycle rider for being a pedestrian or a stop sign. If it thinks that there are two stop signs, it might hit the rider. On the other hand, DOMINO would assign lower penalties for confusing pedestrian and bicycle rider, whereas it assigns a maximal penalty for confusing either of these classes with the stop sign. DOMINO helps reduce the risk to the bicycle rider.

So far, DOMINO has been added to segmentation applications in T1-weighted Magnetic Resonance Images (T1-MRIs) and the Cityscapes dataset [5]. Our results in T1-MRIs are featured in Stolte et al. [17]. Our classification studies have been on the MEDNIST [20], MNIST [13], and FashionMNIST [19] datasets. Some results on MEDNIST are featured in this section to show the promise of our method.

Figure 1 shows our confusion matrix results using a simple convolutional neural network (CNN) with crossentropy 1a, DOMINO-HC 1b, and DOMINO-CM 1c. These results show that our overall accuracy is 99.47% with cross-entropy, 99.54% with DOMINO-HC, and 99.56% with DOMINO-CM. Both of our methods increases the accuracy even further. Both DOMINO-CM and DOMINO-HC also improve segmentation accuracy [17].

Table 1: Brier Score Loss [1] for basic cross-entropy and one of our methods. A lower Brier Score Loss corresponds to better calibration. The best performances are in bold.

| MEDNIST Class | Cross-Entropy | DOMINO-HC | DOMINO-CM |
|---|---|---|---|
| Abdomen CT | 0.00219 | 0.00193 | **0.00183** |
| Breast MRI | 0.145 | 0.144 | **0.143** |
| CXR | 0.163 | **0.160** | 0.161 |
| ChestCT | 0.170 | 0.169 | **0.167** |
| Hand | 0.172 | **0.170** | **0.170** |
| Head CT | 0.162 | 0.162 | **0.160** |

Table 1 shows that these methods also give lower Brier Scores [1] Brier score loss measures the difference between predicted probability and the model’s assigned probability outputs for a given class [1]. A model’s assigned probability outputs refer to the softmax outputs before the final label assignment. A lower Brier score corresponds to better calibration.

## 4. How to implement DOMINO

DOMINO can be used within any PyTorch-based deep learning loss function. Our previous experiments have tested DOMINO on both classification and segmentation. In addition, DOMINO works with different deep learning algorithms. Prior experiments have tested simple CNNs, U-Net transformers (UNETR) [10], and DeepLabv3+ [3]. In these experiments, DOMINO was paired with cross-entropy loss, dice loss, or a combined cross-entropy and dice loss. Therefore, DOMINO's only requirement for functionality is that it currently requires the code to be written in PyTorch.

Computation of the $W$ matrix for a specific user task is required to run DOMINO. One advantage of our previous confusion matrix method is that it requires no previous knowledge of a problem to calculate $W$ [17]. Broadly, this loss penalty can be calculated from the normalized confusion matrix of an uncalibrated model. This approach requires twice the running time of the uncalibrated model or hierarchical DOMINO method. On the other hand, the hierarchical grouping method does require manual creation of the required task groups [17]. This approach requires some knowledge of the intended task output, but under the right circumstances it can leverage user knowledge and intuition to provide improved outcomes. One example would entail grouping disease severity levels or sub-types based on their common treatment recommendations to minimize the impact of the most likely incorrect labeling results. All $W$ constructions require the diagonal to be all zeros, as the diagonal of $W$ represents "the penalty of correct classification".

CUDA GPU access is not required, but it is strongly recommended. DOMINO is substantially faster on classification tasks, as classification tasks require image-wise computation whereas segmentation tasks require pixel-wise computation. Decent GPU resources are particularly recommended for segmentation with the confusion matrix method. Our previous work used an A100 NVIDIA GPU for training volumetric image segmentation, whereas MNIST classification could function on a GPU or CPU.

## 5. The impact of DOMINO on current research questions

Uncalibrated models may look good in research development but be untrustworthy in real-life high-risk applications. Thus, there is a gap "from bench to bedside," or "from lab to life". This paper provides a tool that is easy for researchers to implement in their codes to improve calibration. This will allow different projects to preserve their state-of-the-art performance while being calibrated. Therefore, our project has great potential for improving the trustworthiness and reliability of existing deep learning methods in high-risk applications. Another strength of this approach is that it works for classification and segmentation.

DOMINO will be the most beneficial in deep learning tasks where the penalties of making mistakes among different classes are not equal, especially in high-risk applications like medical treatment (e.g., tumor segmentation [11]), self-driving vehicles [16], and financial decision making [8]. In tumor segmentation, a triage system that recognizes healthy tissue as a tumor lesion would lead to an unnecessary doctor consultation, whereas a system that recognizes a tumor lesion as healthy could cause a person with cancer to miss their critical treatment window. Here, DOMINO would give a larger weight penalty for mistaking tumor tissue as healthy tissue than vice versa. Similarly, a self-driving car would need DOMINO to give a larger penalty for confusing humans for non-humans than the reverse. Further, DOMINO's implementation could lead the respective software to make choices that are oriented towards lower downstream error, rather than focusing on traditional accuracy metrics. For example, the DOMINO creators have published an academic paper on DOMINO for medical image segmentation [17]. A following paper is planned that extends this segmentation to penalize head tissues based on their properties in applications for non-invasive brain stimulation.

In addition to its future impact on high-risk software, DOMINO will benefit ongoing research questions in machine learning calibration. Specifically, multi-class calibration problems are even less well-defined than binary calibration. Typical methods treat classes in a one-vs-all or all-vs-all approach [12]. The area of multiclass calibration needs more significant research attention to develop better evaluation criteria. DOMINO contributes towards this area by challenging the idea that equal class treatment (e.g., in the one-vs-all approach) is reflective of the reliability and trustworthiness of deep learning systems. The research community and many application areas will benefit from this software's contribution to trustworthy machine learning.

The results enabled by the software have been reported in the following academic publication:

Skylar E. Stolte, Kyle Volle, Aprinda Indahlastari, Alejandro Albizu, Adam J. Woods, Kevin Brink, Matthew Hale, and Ruogu Fang. DOMINO: Domain-aware Model Calibration in Medical Image Segmentation. International Conference on Medical Image Computing and Computer Assisted Intervention (MICCAI) 2022. 2022. Conference Proceedings, Springer. DOI: https://doi.org/10.1007/978-3-031-16443-9 44. See [17]

**Declarations of Competing Interests**

The authors declare that they have no known competing financial interests or personal relationships that could have appeared to influence the work reported in this paper.

**Acknowledgements**

This work was supported by the National Institutes of Health / National Institute on Aging (NIA RF1AG071469, NIA R01AG054077), the National Science Foundation (1908299), the Air Force Research Laboratory Munitions Directorate (FA8651-08-D-0108 TO48), and the NSF-AFRL INTERN Supplement (2130885). We acknowledge the people of the NVIDIA AI Technology Center (NVAITC) for their suggestions.

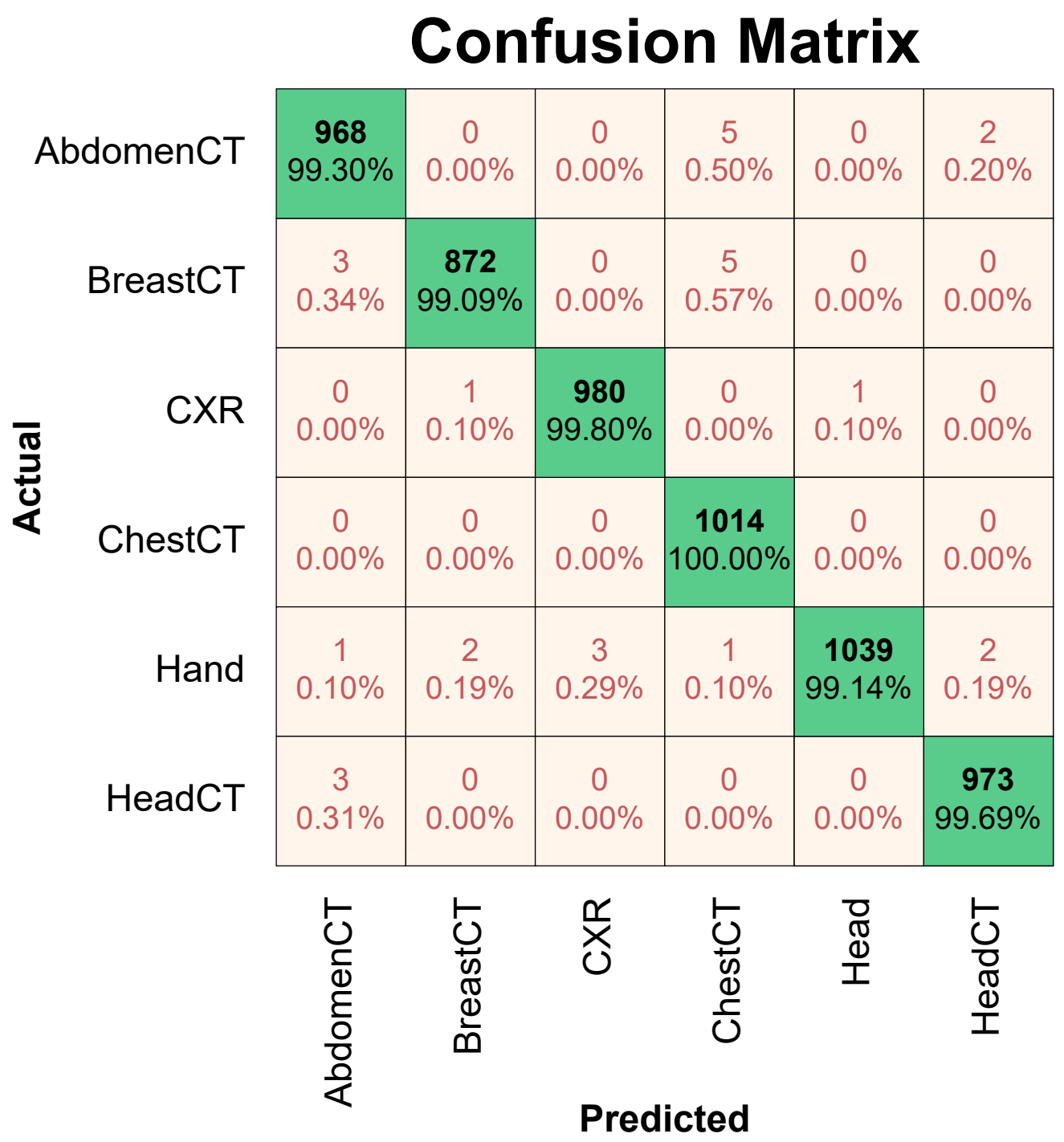

(a)Cross-Entropy (CE)

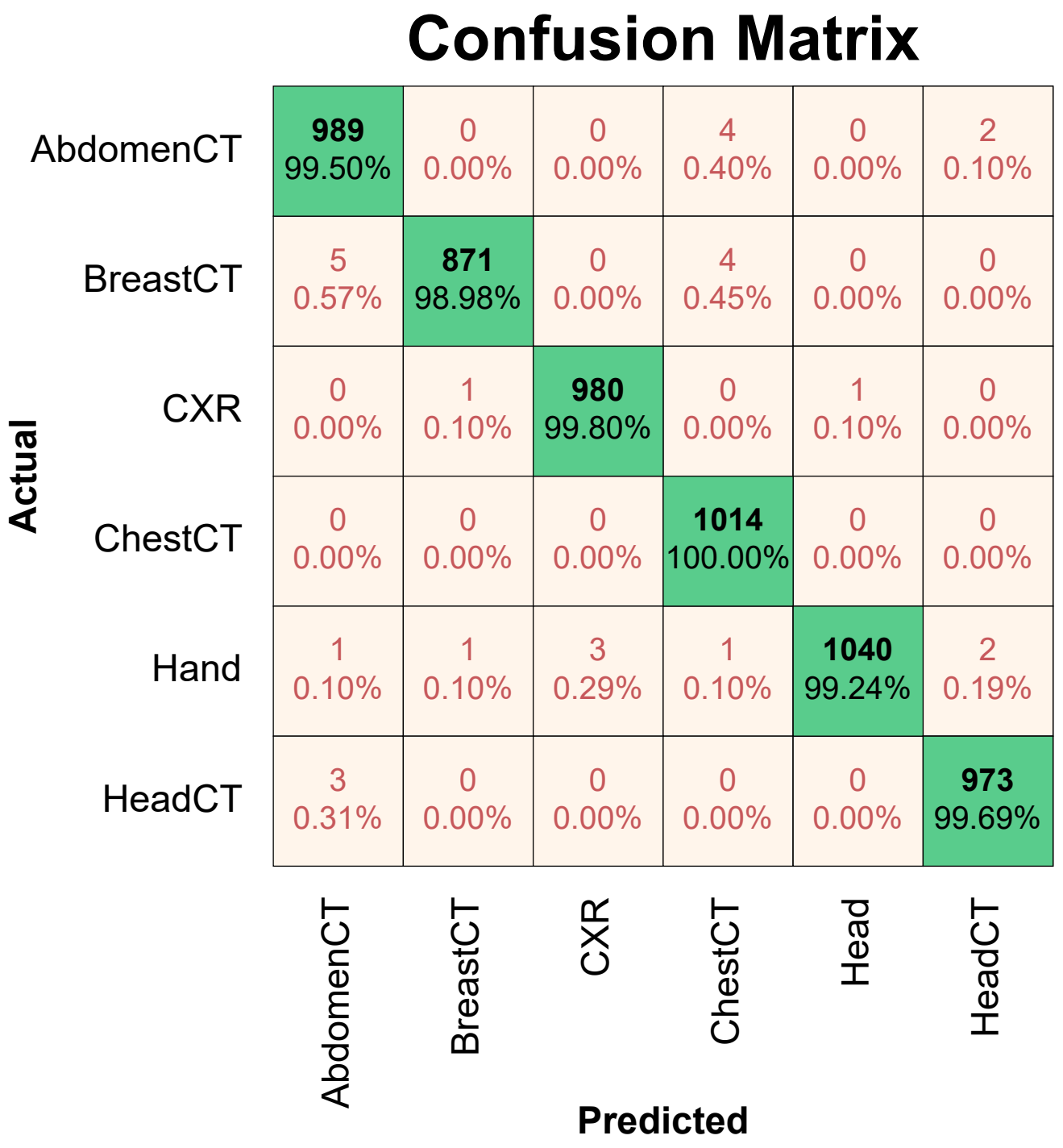

(b)DOMINO-HC

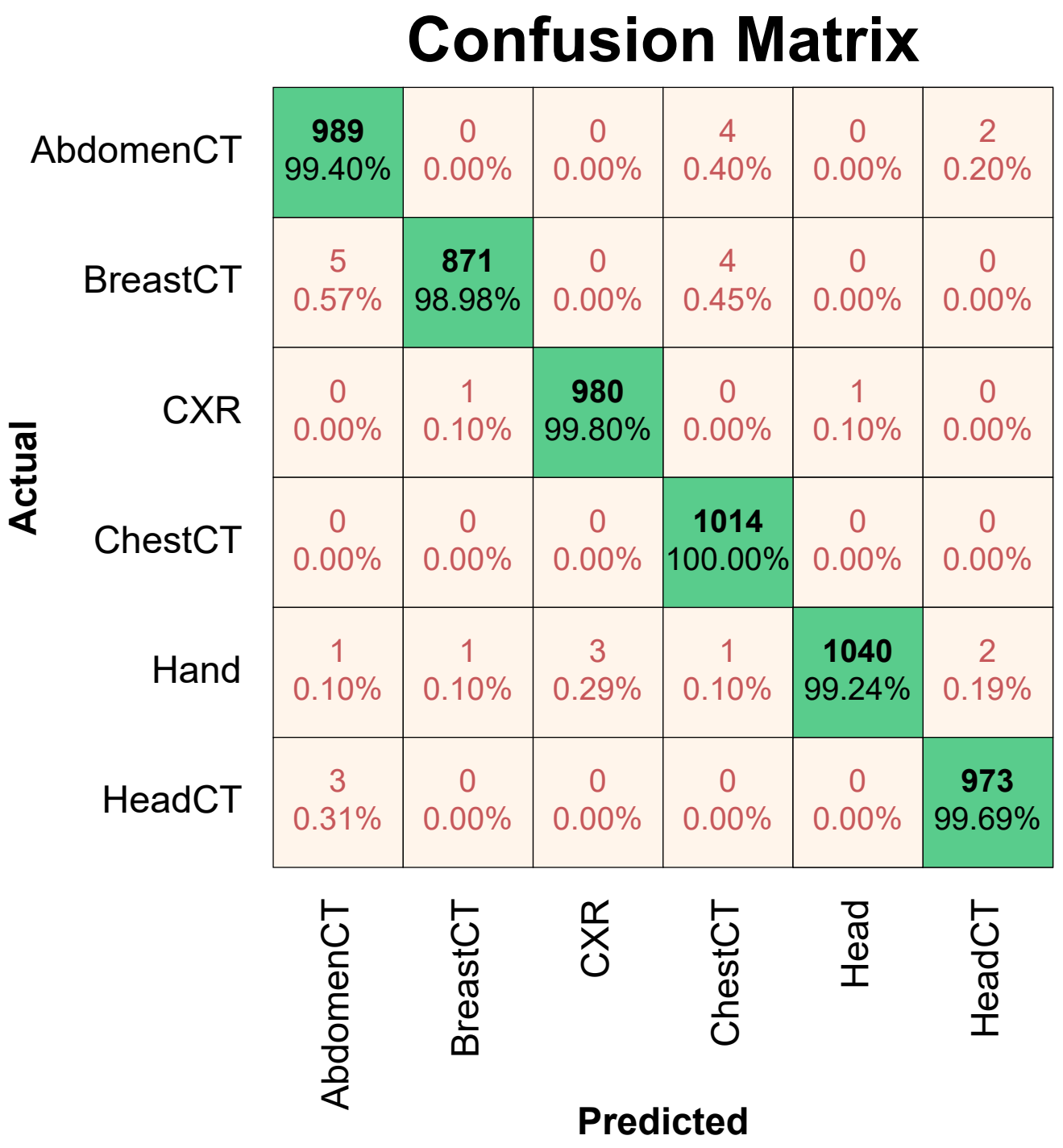

(c)DOMINO-CM

Figure 1: Confusion matrices on testing set of MEDNIST classification. The cross-entropy model gives 99.47% performance, DOMINO-HC gives 99.54% performance, and DOMINO-CM gives 99.56% performance. Notably, DOMINO-HC and DOMINO-CM confuses AbdomenCT for ChestCT and Hand for BreastCT at lower rates.